\pdfoutput=1

\documentclass[11pt]{article}

\usepackage[final]{acl}

\usepackage{times}
\usepackage{latexsym}

\usepackage[T1]{fontenc}

\usepackage[utf8]{inputenc}

\usepackage{microtype}

\usepackage{inconsolata}

\usepackage{graphicx}
\usepackage{subfigure}
\usepackage{subcaption}
\usepackage{hyperref}

\usepackage{newtxtext,newtxmath}
\usepackage{xcolor}
\usepackage{svg}

\title{KodeXv0.1: A Family of State-of-the-Art Financial Large Language Models}

\author{
 \textbf{Neel Rajani\textsuperscript{1,2}},
 \textbf{Lilli Kiessling\textsuperscript{1,3}},
 \textbf{Aleksandr Ogaltsov\textsuperscript{1}},
 \textbf{Claus Lang\textsuperscript{1}},
\\
 \textsuperscript{1}{\protect\includegraphics[ trim=0 5cm 0 0, width=0.7cm]{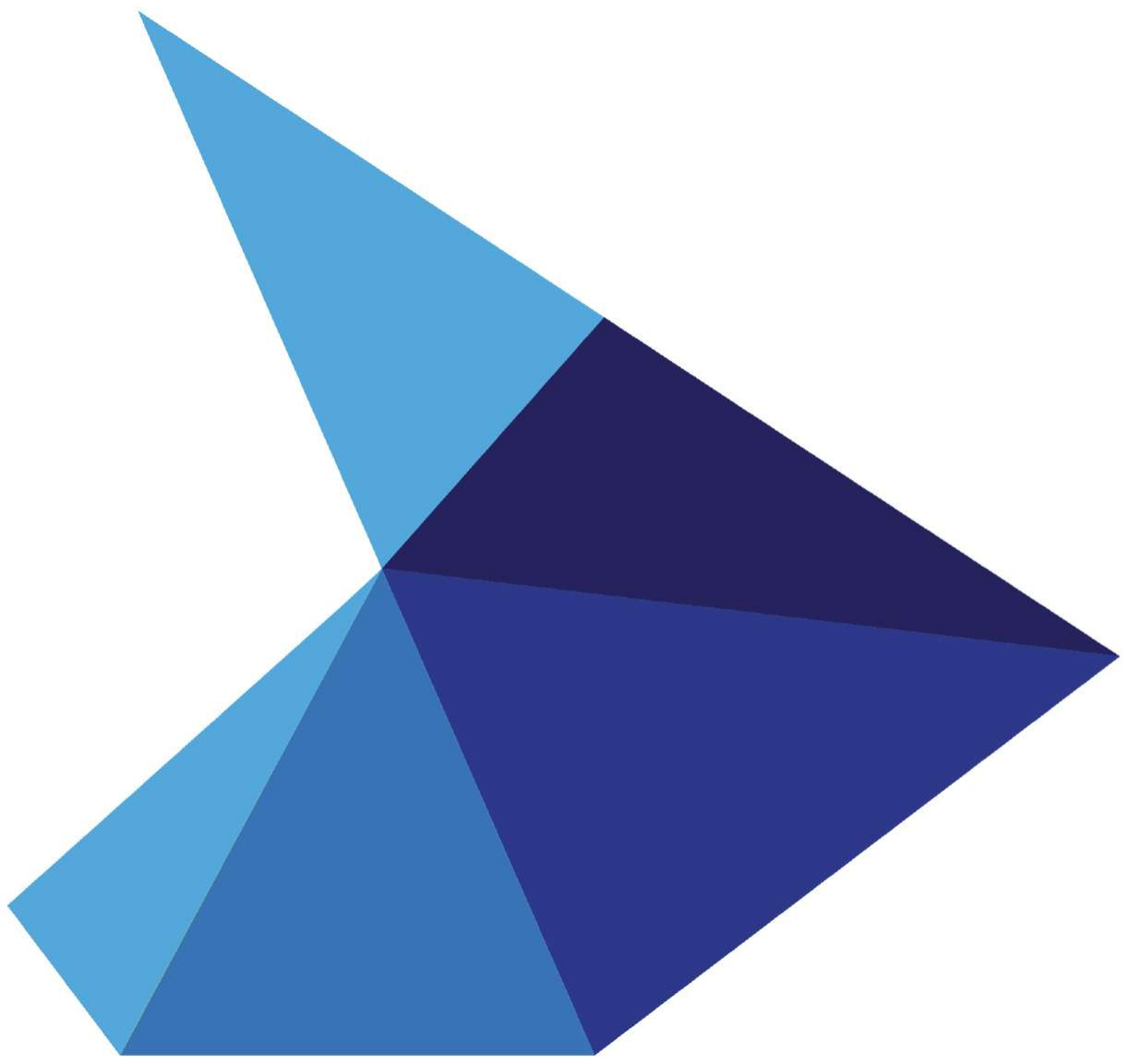}}KodexAI,
 \textsuperscript{2}University of Edinburgh,
 \textsuperscript{3}BCCN Bernstein Center for Computational Neuroscience,
\\
 \small{
   \href{mailto:email@domain}{Neel.R@web.de},
   \href{mailto:email@domain}{lillikiessling@gmail.com},
   \{claus.lang, aleksandr.ogaltsov\}@kodex-ai.com
 }
}

\begin{document}
\maketitle
\begin{abstract}
Although powerful, current cutting-edge LLMs may not fulfil the needs of highly specialised sectors. We introduce KodeXv0.1, a family of large language models that outclass GPT-4 in financial question answering. We utilise the base variants of Llama 3.1 8B and 70B and adapt them to the financial domain through a custom training regime. To this end, we collect and process a large number of publicly available financial documents such as earnings calls and business reports. These are used to generate a high-quality, synthetic dataset consisting of Context-Question-Answer triplets which closely mirror real-world financial tasks. Using the train split of this dataset, we perform RAG-aware 4bit LoRA instruction tuning runs of Llama 3.1 base variants to produce KodeX-8Bv0.1 and KodeX-70Bv0.1. We then complete extensive model evaluations using FinanceBench, FinQABench and the withheld test split of our dataset. Our results show that KodeX-8Bv0.1 is more reliable in financial contexts than cutting-edge instruct models in the same parameter regime, surpassing them by up to 9.24\%. In addition, it is even capable of outperforming state-of-the-art proprietary models such as GPT-4 by up to 7.07\%. KodeX-70Bv0.1 represents a further improvement upon this, exceeding GPT-4's performance on every tested benchmark.
\end{abstract}

\section{Introduction}

Large language models (LLMs) have seen increasing application in the financial sector, ranging from automated report generation and sentiment analysis to risk assessment and compliance monitoring \citep{caoli2024}. The ability of LLMs to process complex documents and generate human-like text has positioned them as valuable tools for decision-making processes and improving operational efficiencies within financial institutions. However, despite their strengths, several challenges remain before LLMs can responsibly be deployed. Financial language is characterized by specialized terminology and a high degree of contextual dependence. Moreover, business documents are frequently multilingual, reflecting the global nature of financial markets. Cutting-edge proprietary models often struggle to generate accurate responses in contexts that require a deep understanding of financial concepts, and submitting sensitive information to their APIs may infringe upon privacy regulations. These challenges underscore the need for specialized fine-tuning and data curation strategies to develop LLMs that are better adapted to highly specialized sectors. In this work, we present a family of LLMs designed to perform more reliably across various financial NLP tasks. Our report outlines the methodologies we used to exceed the performance of proprietary models in the financial domain. 

\begin{figure}[t]
    \centering
    \includegraphics[width=0.95\linewidth]{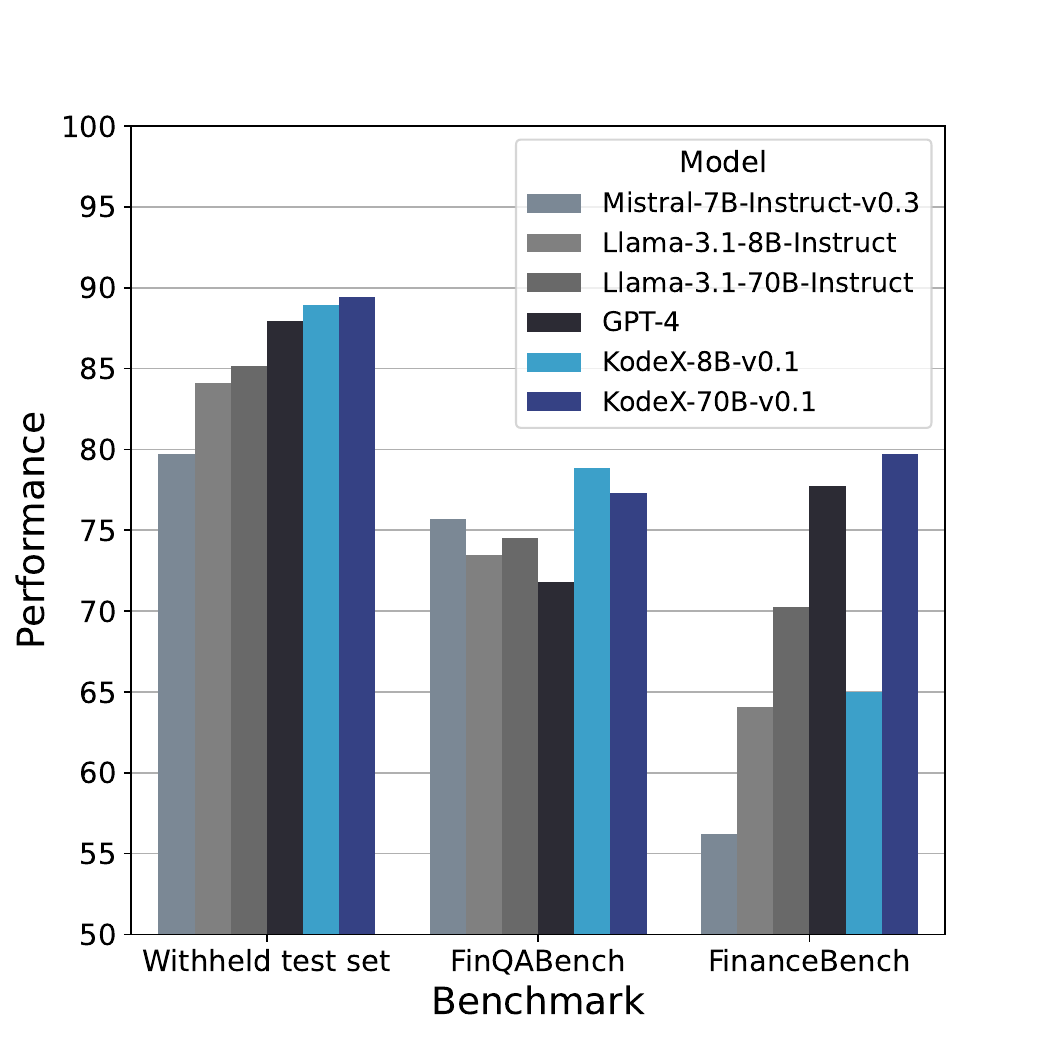}
    \caption{Performance comparison between KodeXv0.1 models against open-source instruct models and GPT-4. The 8B variant exhibits best-in-class performance, and the 70B variant outmatches GPT-4 on every benchmark.}
    \label{fig:main_results_graph}
\end{figure}

\begin{figure*}[h]
    \centering
        \includegraphics[width=\textwidth]{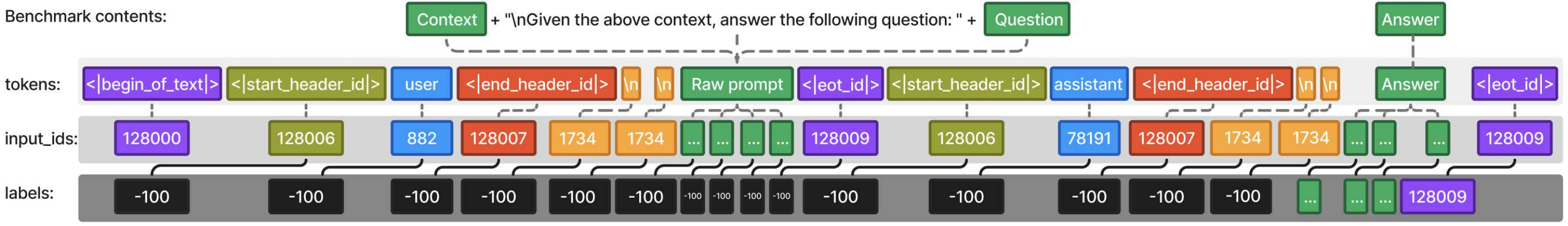}
        \caption{\label{fig:instruction-masked fine-tuning} Our instruction-tuning setup. A benchmark instance is parsed, inserted into the Llama 3.1 instruction-tuning format, and tokenized. The labels are created by shifting the inputs to the left by one, and masking tokens from the instruction with -100. This signals to PyTorch that these labels should be ignored when computing the loss.}
\end{figure*}

\section{Related Works}

\subsection{Financial LLMs and their evaluation}  
Adapting language models to specific domains has emerged as a promising path towards enabling real productivity gains. A substantial contribution towards this area of research was made with BloombergGPT by \citet{wu2023bloomberggptlargelanguagemodel}, who showed that mixing general-purpose and domain-specific data can be lucrative for financial contexts. Similarly, \citet{bhatia2024fintralfamilygpt4level} apply recent innovations in fine-tuning methodologies to Mistral-7Bv0.1. We further build on these findings by using the latest open-source models from the Llama 3.1 family.

The financial capabilities of such LLMs are typically measured through domain-specific benchmarks. An example of this that has seen wide adoption is FinQA by \citet{chen-etal-2021-finqa}, a large-scale dataset with question-answer pairs on financial reports, written by industry experts. However, the benchmark may not closely model real-world contexts as it focuses on abstract numerical reasoning in the form of reasoning programs, instead of text generation using financial language with qualitative factors present in real-life situations. An extension upon FinQA is built by \citet{chen2022convfinqaexploringchainnumerical}, who transform question instances by simulating a corresponding conversation flow to produce ConvFinQA. Nonetheless, the ground truth per question often only consists of a singular numeric response, and in the financial context, a long-form response may be more desirable. This is the case in FinanceBench by \citet{islam2023financebenchnewbenchmarkfinancial}, who employ a multidisciplinary team of experts to create and annotate context-question-answer triplets. Similarly, \citet{AI_2024a} created a financial context-question-answer benchmark in FinQABench that is manually verified by human annotators.

\subsection{Instruction following}
Since it was popularized by the InstructGPT paper \citep{ouyang2022traininglanguagemodelsfollow}, fine-tuning for instruction following has become an essential phase towards the end of training for any cutting-edge LLM \citep{jiang2023mistral7b, dubey2024llama3herdmodels, yang2024qwen2technicalreport}. After being initially trained on vast amounts of Internet data, LLMs can sometimes generate unfiltered outputs. To address this, a second phase of fine-tuning (focused on instruction following) is necessary to adapt the model for controlled use as a chat bot or assistant. The Self-Instruct paper and the Alpaca paper by \citet{wang2023selfinstructaligninglanguagemodels} and \citet{alpaca}, respectively, demonstrated the feasibility of aligning smaller, open-source models on synthetic data too. Since, the alignment of LLMs into assistants via instruction following has become standard practice, but there is nuance to how it is achieved. Specifically, a recent paper by \citet{shi2024instructiontuninglossinstructions} emphasizes the distinction between \textit{instruction modelling} and \textit{instruction tuning}. In instruction modelling, the "label" or "target" for each token is simply the next token. The cross-entropy loss is computed for the prediction of every token in the instruction \textit{and} the response. In contrast, during instruction tuning, the loss is only computed on tokens in the response, as illustrated in Figure \ref{fig:instruction-masked fine-tuning}. How these approaches behave differently in financial contexts, however, remains under-researched.

\section{Methods and Experimental Setup}
\subsection{Training data generation}

The impact of dataset selection for LLM training has been shown repeatedly to have a pronounced impact on downstream performance \citep{ivison2023camelschangingclimateenhancing, groeneveld2024olmoacceleratingsciencelanguage, biderman2023pythiasuiteanalyzinglarge, dubey2024llama3herdmodels}.  Recently, the use of synthetic datasets has emerged as a reliable way to train well-performing LLMs \citep{puri-etal-2020-training}. Notably, \citet{dubey2024llama3herdmodels} highlight the potential of using synthetic data during the post-training stage that they apply to Llama 3.1 base models. As such, we choose to curate a high-quality, synthetic dataset based on financial documents. These were obtained through systematic scraping and parsing of publicly available data from prominent companies listed on major indices such as the MSCI World and European indices. A total of 822 documents from over 40 companies were utilized to generate the training dataset. We employ a multi-stage dataset generation pipeline to use these documents as the source for our synthetic data.

\begin{figure}
        \centering
        \includegraphics[width=0.5\textwidth]{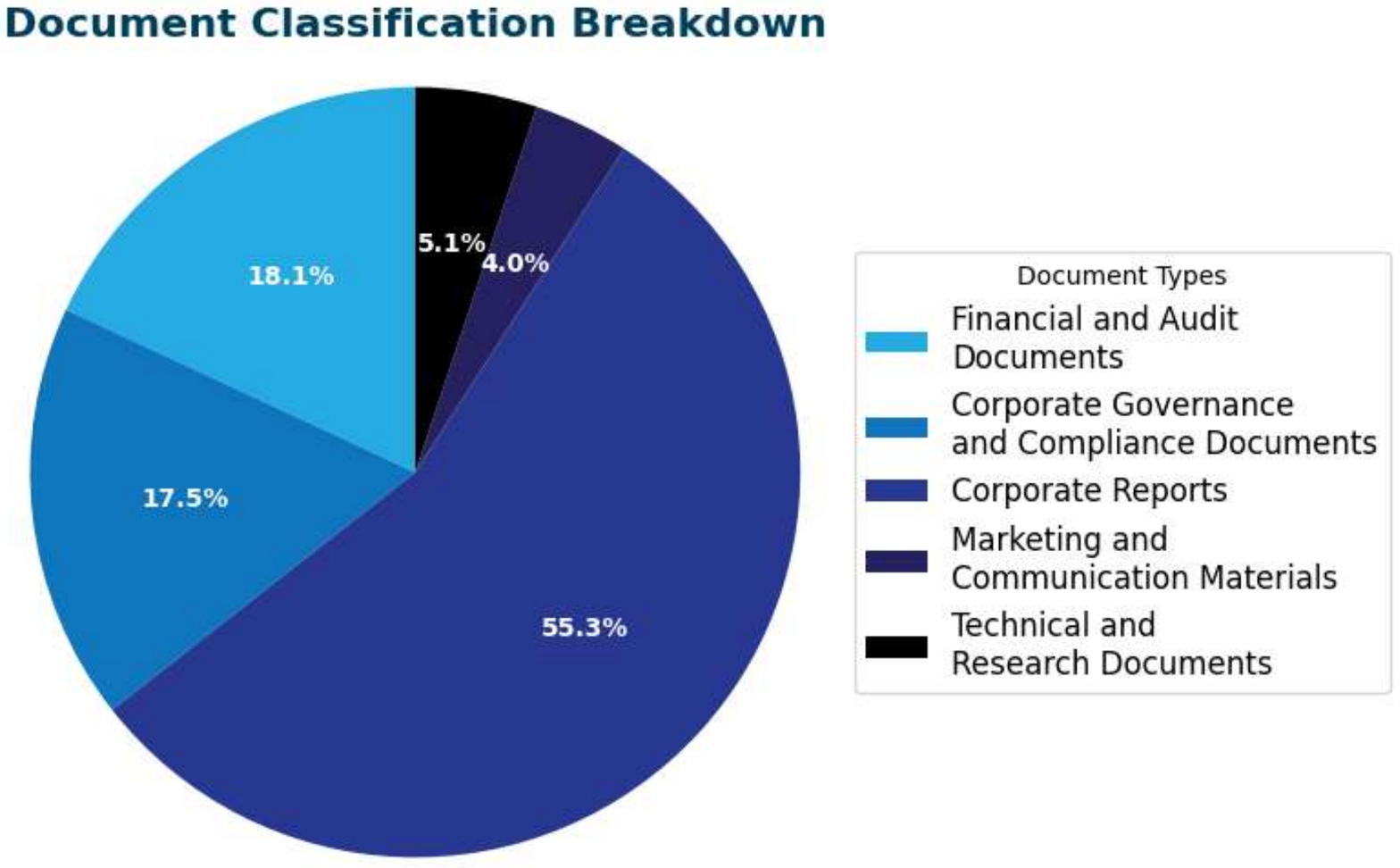}
        \caption{ Distribution of categories from which the documents in the training data were chosen. }
        \label{fig:categories-training-data}
\end{figure}

The documents were predominantly in English, though a significant portion included various other languages. As such, our dataset comprises over 20\% multilingual content in languages such as German, French, Norwegian, Swedish, Spanish, Italian, and more. This helps enhance the model's performance in multilingual tasks, which is particularly relevant in financial sector communications. Furthermore, the training documents encompass a broad spectrum of categories, including Financial and Audit Documents, Corporate Governance and Compliance Documents (e.g., Codes of Ethics and Conduct), Legal Documents, Corporate Reports (such as Annual Reports, Equity Research, Corporate Social Responsibility Reports, Sustainability Reports, and Government Reports), Marketing and Communication Materials, as well as Technical and Research Documents. The distribution of these document categories is illustrated in Figure \ref{fig:categories-training-data}. 

\begin{figure}
    \includegraphics[width=0.99\linewidth]{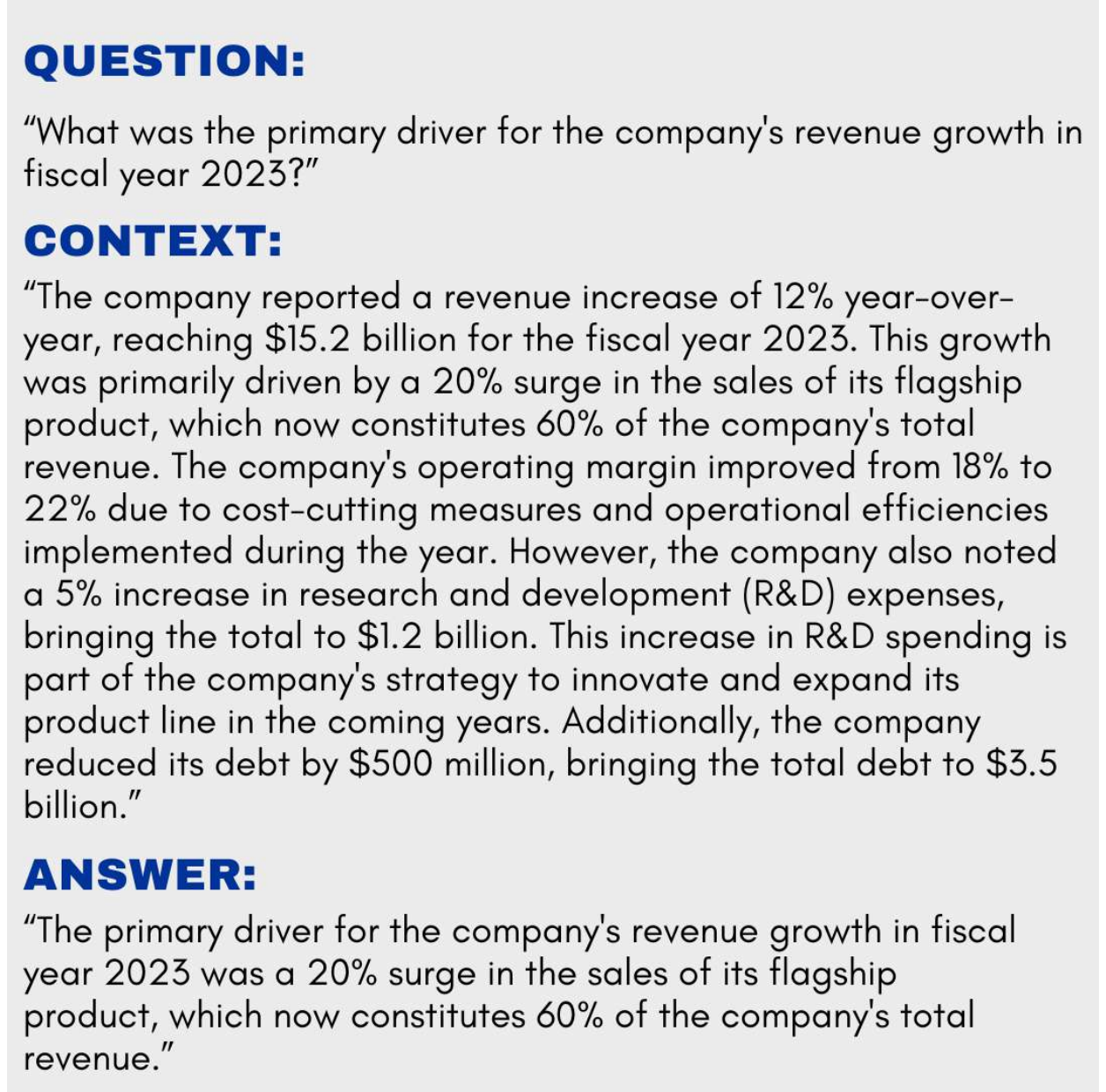}
    \caption{Example of one training sample.}
    \label{fig:training-sample}
\end{figure}

After collection and processing, the documents were segmented into context chunks, and question-answer pairs were synthesized based on the provided context. In total, we generated 48,415 samples, which were then randomized and separated into a 47,415-sample train split and a 1,000-sample test split. The train split comprises a total of 54,508,439 tokens when using the Llama 3.1 tokenizer and an example of a training sample is depicted in Figure \ref{fig:training-sample}.

Additionally, the dataset includes questions that are not answerable from the context. In those instances, the corresponding reference answer explicitly states that the question is unanswerable. By training on a dataset that includes such instances, we aim to teach the model not to confine generations to always extract some kind of answer from the context, because in cases where this is not possible, hallucinations are widespread. We observe that this is an essential step towards reducing the occurrence of hallucinations in the fine-tuned model, in line with previous research \cite{rajpurkar-etal-2018-know}.

\subsection{Benchmarking}  
\label{section:benchmarking}

During the evaluation of our fine-tuning efforts, we turn towards the following three benchmarks to gain a nuanced understanding of model behaviour.
\begin{itemize}
    \item \textbf{Withheld test set:} The test set is a diverse collection of 1,000 questions from the synthetic dataset, covering a wide range of companies, languages and financial contexts. Performance on this test set is the main signal for our decision-making on all factors relating to training, such as hyperparameters, total steps and epochs, base model and others. We opt for this approach as it gives us a direct sanity check of whether a model is learning from the training set, and to be able to measure in-domain performance. 
    \item \textbf{FinanceBench:} FinanceBench was created by \citet{islam2023financebenchnewbenchmarkfinancial} and comprises questions on provided context relating to publicly traded companies. It provides a variety of question types, such as information extraction, logical reasoning and numerical reasoning. 
    \item \textbf{FinQABench:} FinQABench was created by \citet{AI_2024a} for the evaluation of hallucination and accuracy of LLMs in the financial domain. We find minor errors in the test set and hence choose to verify and correct all questions manually.
\end{itemize}

Both of the latter benchmarks involved human experts during creation and annotation, which is a crucial element to bolster the validity of our results outside of the training distribution, in addition to our in-distribution synthetic test set.

As all three benchmarks feature a Context-Question-Answer setup, we employ the same prompting strategy for every benchmark, shown in Appendix \ref{sec:prompts}. The generated answers are then marked in comparison to the reference answers. While human evaluation is the gold standard for assessing model performance, it is prohibitively slow and costly. To enable us to iterate quickly, we make use of proprietary LLMs, such as Gemini, in an LLM-as-a-judge setting. \citet{zheng-etal-2024-judging} find that LLM judgements match human preferences very well and that Chain-of-thought and reference-guided judging improves reliability. We follow these guidelines in designing our evaluation prompt, also shown in Appendix \ref{sec:prompts}. The judge assigns a score between 1 and 10 to each generated answer, based on its alignment with the reference answer. The scoring allows for the quantitative evaluation of the fine-tuned model's performance and for comparison with available base models. Inspired by \citet{wang2023selfconsistencyimproveschainthought}, each question is judged three times and the average is recorded to improve the robustness of judgments. 

\subsection{Training details}

We choose Llama 3.1 as our base model due to its cutting-edge performance, permissive license, and availability in different sizes and variants. In preliminary trials, we experiment with the use of base and instruct variants of Llama 3.1. Notably, we observe that while instruct variants exhibit much higher performance without training, the base variants can surpass these scores after training, as shown in Figure \ref{fig:main_results_graph}. Therefore, we adopt the established methodology of the Meta AI team behind Llama 3 \citep{dubey2024llama3herdmodels} and train the Llama 3.1 base models for instruction following using a synthetic dataset in the post-training phase. However, our approach diverges by focusing on domain-specific data from the financial sector only. 

In this phase, we experiment with both instruction modelling and instruction tuning. Results 
by \citet{shi2024instructiontuninglossinstructions} suggest using instruction modelling for datasets with long prompts and short responses. They argue that by forcing the model to improve at predicting the instruction too, it learns more about the target domain. However, \citet{huertaenochian2024instructionfinetuningdoesprompt} instead find that including loss on instructions when using datasets with short responses degrades performance. We similarly find that instruction modelling is consistently outperformed by instruction tuning. This may be related to the nature of our dataset: The context field per instance comes directly from financial documents, which include elements such as tables, table descriptions and formatting information. We want to train our models to improve not at predicting such context tokens and question tokens, but instead make use of them to correctly predict answer tokens. As such, we carefully overwrite labels of the instruction with [-100] when tokenizing training samples, as shown in Figure \ref{fig:instruction-masked fine-tuning}.  

The Llama 3.1 base models are loaded with 4bit quantization using bitsandbytes \citep{dettmers2022llm}, and wrapped for LoRA training \citep{hu2021loralowrankadaptationlarge} using PEFT \cite{peft}. To ensure stable and efficient training, we make use of the TRL library \citep{von_Werra_TRL_Transformer_Reinforcement}. Observing performance degradation in multi-epoch training, we train our KodeXv0.1 models for 1 epoch only. To speed up training, we make use of various recent innovations, such as Flash Attention 2 \citep{dao2023flashattention2fasterattentionbetter} and torch.compile() by the PyTorch team \citep{Ansel_PyTorch_2_Faster_2024}.
To monitor training runs and perform principled hyperparameter sweeps, we employ WeightsAndBiases \citep{wandb}. The results of our sweeps of LoRA parameters and training hyperparameters are detailed in Appendix \ref{sec:hyperparams}.

\section{Results and Evaluation}

\begin{table*}[]
\centering
\begin{tabular}{lllll}
\hline
\textbf{Model}           & \textbf{Size} & \textbf{Withheld test set} & \textbf{FinQAbench} & \textbf{FinanceBench} \\ \hline
Mistral-7B-Instruct-v0.3 & 7B            & 79.69                      & 75.67               & 56.24                 \\
Llama-3.1-8B-Instruct    & 8B            & 84.11                      & 73.50               & 64.09                 \\
Llama-3.1-70B-Instruct   & 70B           & 85.18                      & 74.50               & 70.24                 \\ \hline
GPT-4                     & 1.6T*         & 87.95                      & 71.80               & 77.76                 \\ \hline
KodeX-8Bv0.1             & 8B            & 88.96                      & \textbf{78.87}      & 64.98                 \\
KodeX-70Bv0.1            & 70B           & \textbf{89.44}             & 77.33               & \textbf{79.69}        \\ \hline
\end{tabular}

\caption{Our main benchmark results. The highest respective performance per benchmark is in \textbf{bold}. Our KodeXv0.1 models set the state of the art in all benchmarks that we tested. *Note that this figure has been the topic of widespread speculation but is yet to be confirmed.}
\label{table:main_results}
\end{table*}

Our main results are shown in Table \ref{table:main_results}. Note that all scores refer to the judgements made in our evaluation phase, as outlined in Section \ref{section:benchmarking}. All experiments were performed with a temperature of 0 and greedy decoding instead of sampling. The deterministic nature of these results eliminates the need to consider variance.

\subsection{Withheld test set}
\begin{figure}[h]
    \centering
    \includegraphics[width=0.95\linewidth]{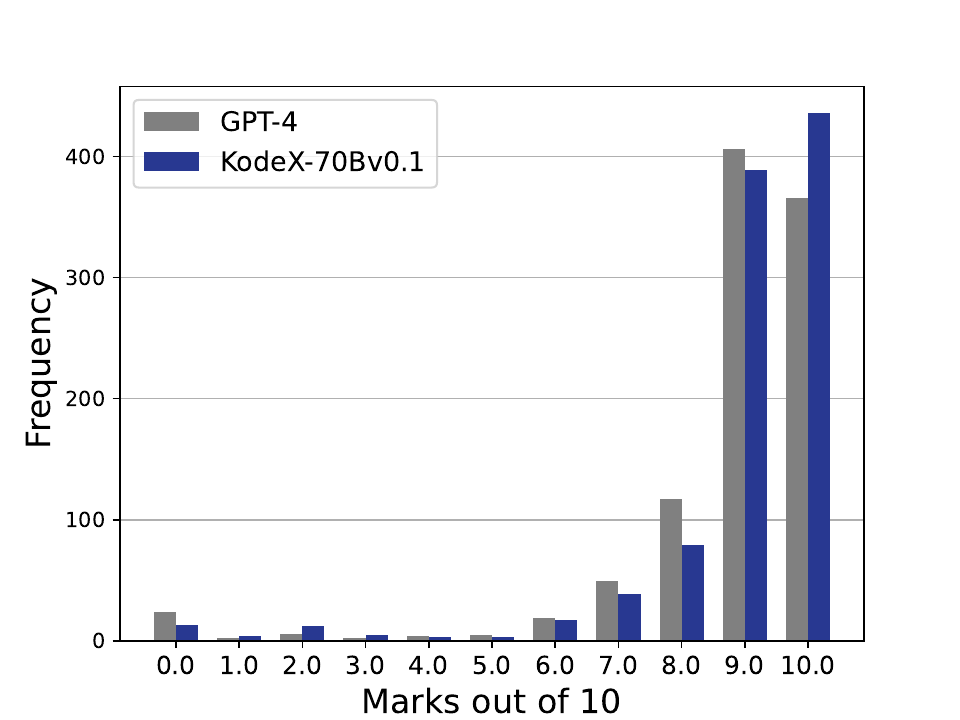}
    \caption{Comparison of the frequency with which KodeX-70Bv0.1 on GPT-4 obtains a certain mark out of 10 on the withheld test set. \textbf{}}
    \label{fig:comparison}
\end{figure}

We find that our training regime allows KodeXv0.1 models to outperform all other cutting-edge models that we benchmarked on the withheld test set. The score of KodeX-8Bv0.1 exceeds that of instruct variants of Mistral and Llama 3.1 in the same parameter regime by up to 9.27 percentage points, and exceeds that of Llama 3.1-70B-instruct and GPT-4 by 3.78\% and 1.01\%, respectively. KodeX-70Bv0.1 surpasses all other models, with a score of 89.44\%. Digging deeper, we can compare how often it achieves a given grade out of 10 compared to GPT-4, as shown in Figure \ref{fig:comparison}.

On the withheld test set, KodeX-70Bv0.1 is less likely than GPT-4 to produce a good to near-perfect response (with scores of 7-9/10), and instead more frequently answers the question fully (with a score of 10/10). This suggests that our KodeXv0.1 models are less likely to recite irrelevant information from the context, and more often include specifically the details that the question asks for. Furthermore, KodeXv0.1 models are less likely to catastrophically hallucinate (with a score of 0/10) than GPT-4. This shows that they are better adapted to financial contexts and process information from documents more faithfully, with high in-distribution performance.

\subsection{FinQABench}
On FinQABench, we similarly observe that KodeX models outperform all other models consistently. Anomalously, we find GPT-4 to attain the lowest score on this benchmark. When looking at outputs and judgements manually, we do observe that GPT-4 struggles with some questions. However, it also occasionally produces lengthy answers that discuss appropriate details of the answer and is then punished for this during LLM-as-a-judge evaluation. This points towards a potential limitation of our investigation. Nevertheless, we note that KodeX models more often answer accurately, especially on numerical and reasoning questions.

\subsection{FinanceBench}
On FinanceBench, we find that Kodex-8Bv0.1 again outperforms models in the same parameter regimen. Notably, the smaller models generally do not perform well on this benchmark. Llama 3.1 70B Instruct and GPT-4 achieve higher scores, especially on complex questions that involve multi-hop reasoning or incorporate multiple concepts. This highlights the difficulty of the benchmark and stands in line with the emerging trend that larger models exhibit greater reasoning capabilities \citep{radford2019language, brown2020languagemodelsfewshotlearners}. However, the larger models are also surpassed by KodeX-70Bv0.1, which even outperforms GPT-4 by 1.93 percentage points.

\section{Conclusion}
We have presented KodeXv0.1, a family of models that advances the cutting edge of LLMs adapted to financial contexts. Our work highlights that proprietary and expensive generalist models can be outpaced by fine-tuning much smaller open-source models on high-quality synthetic data. KodeX-8Bv0.1 comprises a best-in-class model that strikes an attractive balance between its high performance and low memory footprint due to its 4-bit quantisation. This makes it more sustainable and cost-efficient to run, with greater opportunities for local deployment and lower latencies. A further improvement is made by KodeX-70Bv0.1, which outperforms GPT-4 on every tested benchmark and exhibits stellar financial understanding. When placed at the core of our complex production pipeline, our models further benefit from multi-stage RAG architectures, content filters and use-case-specific prompt templating. As such, the performance recorded in this technical report is a starting point for what we envision for our KodeXv0.1 models. They represent a milestone in how we advance data processing and document understanding in the financial domain.

\section{Future research directions}
Given that our KodeXv0.1 models are intended to sit at the heart of our platform, we now look towards evaluating our end-to-end pipeline which we expect to further enhance performance due to our bespoke RAG, content filter and prompt templating modules. Meanwhile, although our synthetic dataset comprises a step towards aligning the training phase of LLMs with real-world contexts, we also note that larger datasets with even greater background diversity and multilinguality may further improve downstream performance. Similarly, we believe there is still headroom in our synthetic data generation, specifically in relation to experimenting with mixes of questions relating explicitly to numerical challenges, multi-hop reasoning and niche financial contexts. Other promising research directions we are excited to pursue include multi-stage training recipes that include alignment phases such as DPO \citep{rafailov2024directpreferenceoptimizationlanguage} or ORPO \citep{hong2024orpomonolithicpreferenceoptimization}. For example, we could envision using models of the largest parameter regimen available such as Llama 3.1 405B or NVIDIA Nemotron-4 340B \citep{nvidia2024nemotron4340btechnicalreport}, which feature permissive licenses that explicitly encourage synthetic data generation. These could be used to create a dataset of "chosen" responses, with responses generated by current KodeXv0.1 models as "rejected" responses, to align our models even further towards desired writing styles.
In a similar vein, the training focus of v0.1 of our models was to advance the state-of-the-art in textual understanding and question-answering of LLMs in financial contexts. However, developing multi-modal models capable of complex tool use comprises another important extension that we are excited to tackle next. 

\section{Limitations}
Despite our efforts to rigorously assess model performance,
our fine-tuning and evaluation pipeline are realistically not free from flaws. For example, we acknowledge that the synthetic nature of our synthetic test set implies that several biases are plausible. Furthermore, in spite of manual verification, it is possible that the benchmark contains irrelevant questions or incorrect answers.  
Similarly, the use of LLM-as-a-judge was a crucial element towards fast iteration and scalability of the training effort. However, it can introduce a number of biases, identified by \citet{zheng-etal-2024-judging} as verbosity bias, self-enhancement bias, and limited capability in grading math and reasoning questions. We carefully design our benchmarking pipeline, judging prompt, choice of judge and question domain to help mitigate these biases to an extent. Nevertheless, it is important to acknowledge that LLM-judged benchmark scores can at best be regarded as a proxy for model performance, and may be obscuring actual capabilities. Further, the training data of the used judge may impact the kind of responses and writing style it prefers. While human judgements also suffer from biases that are hard to eliminate, the biases in LLM-as-a-judge have not been investigated as extensively due to the recent development of the approach. A further bolstering of our results could be achieved through a mixture of evaluations by different human experts and various LLMs. Such an effort goes beyond the scope of this paper but constitutes an interesting subject for subsequent research.

\bibliography{anthology,custom}

\appendix

\section{Inference and evaluation prompts}
\label{sec:prompts}

\begin{figure}[h]
    \centering
    \includegraphics[width=0.95\linewidth]{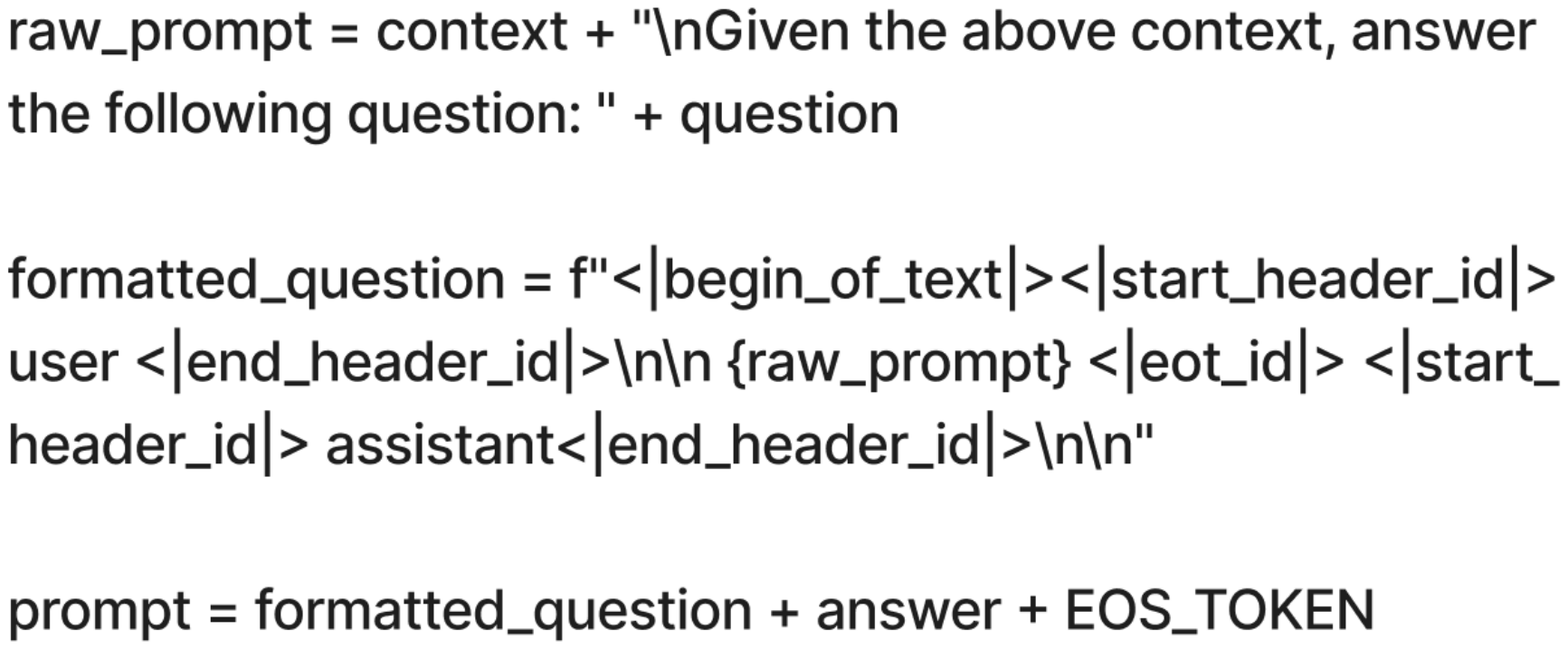}
    \caption{The prompt template used for obtaining responses to benchmark questions from models. This figure includes the instruction template specific to Llama 3.1 models. For other models, the respectively recommended instruction template was used.}
    \label{fig:inference_prompt}
\end{figure}

\begin{figure}[h]
    \centering
    \includegraphics[width=0.95\linewidth]{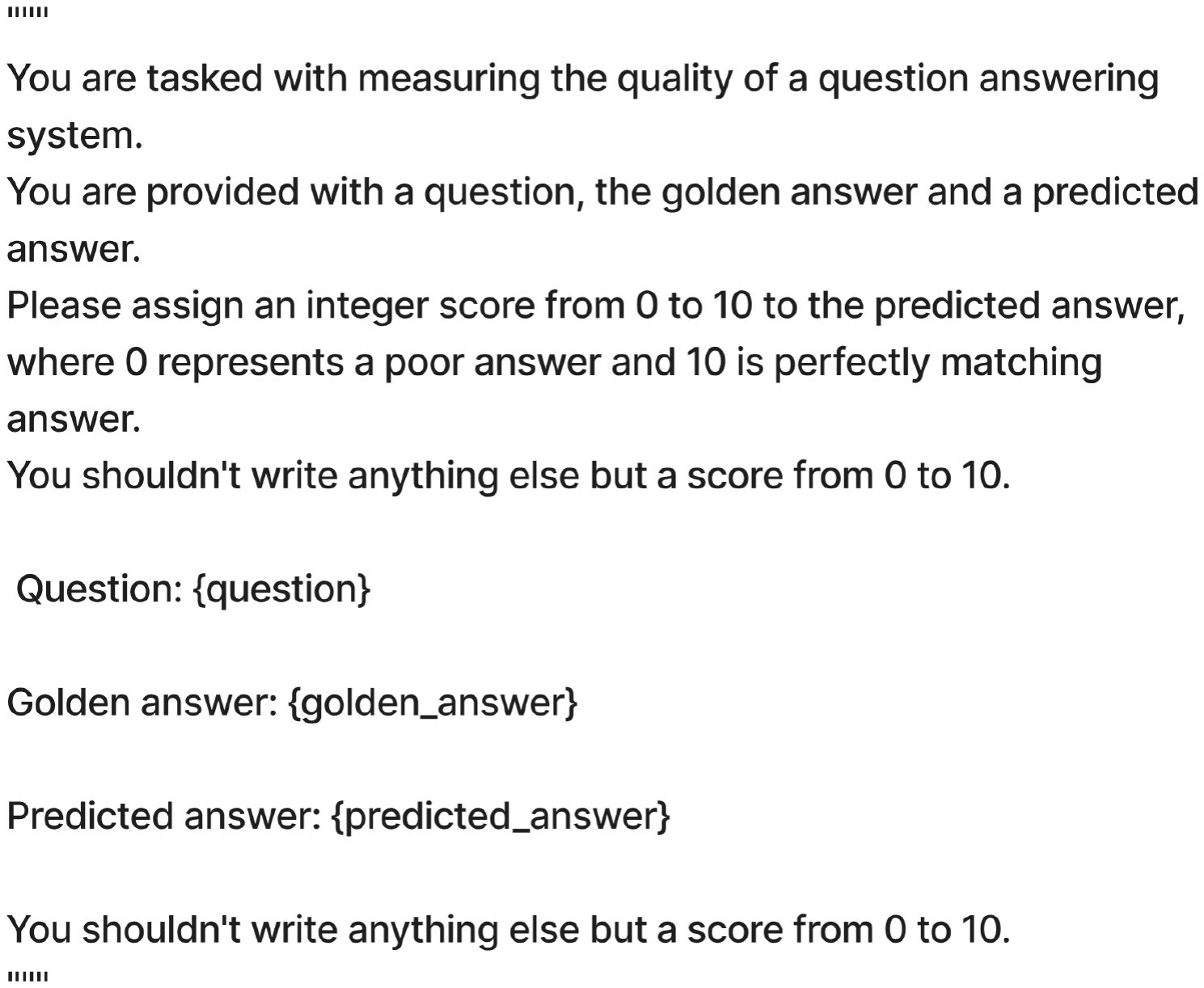}
    \caption{The evaluation prompt. Note the emphasis on generating only a score, which helped output consistency.}
    \label{fig:eval_prompt}
\end{figure}

\section{Hyperparameter choices}
\label{sec:hyperparams}

\begin{figure*}[t]
    \centering
    \includegraphics[width=0.95\linewidth]{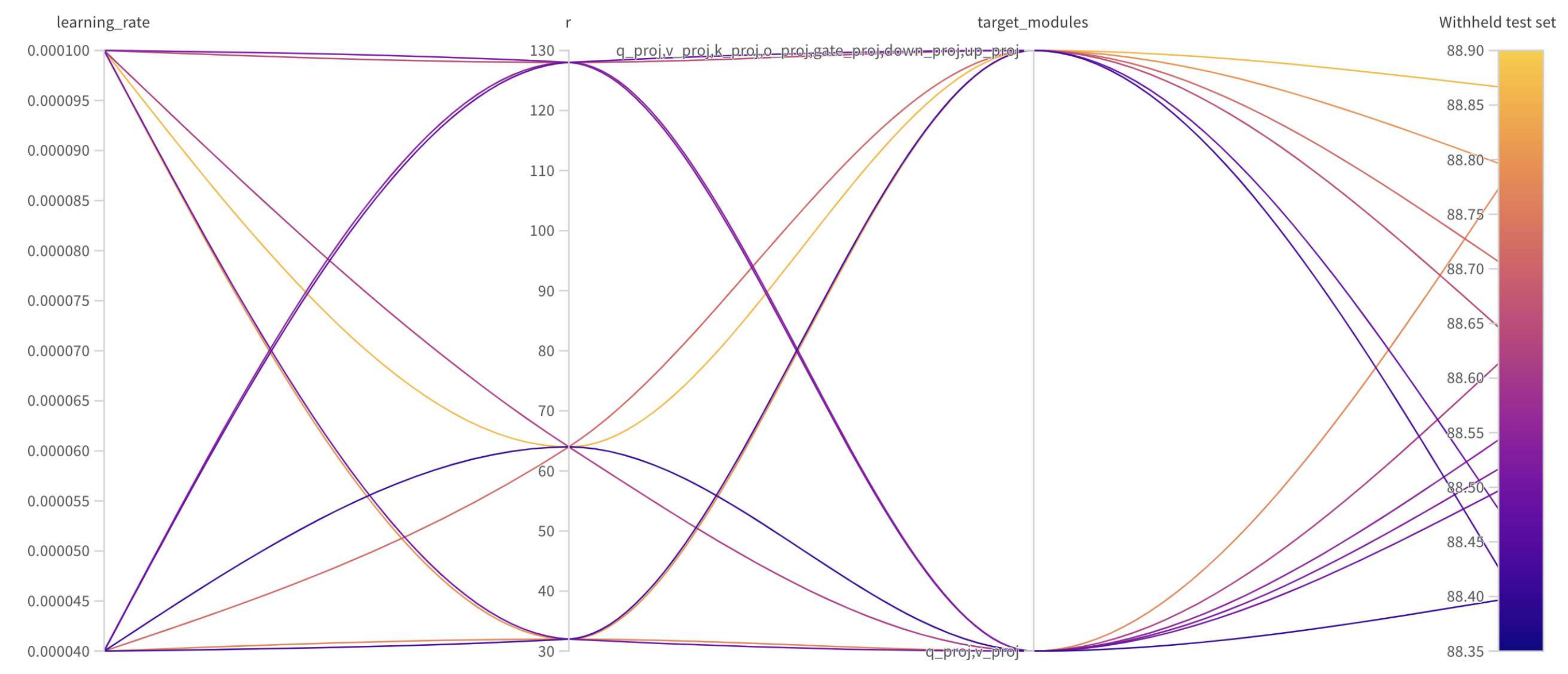}
    \caption{A graph created using WandB by \citet{wandb} detailing the impact of varying learning rates, LoRA ranks and target modules of a KodeXv0.1 8B variant during training.}
    \label{fig:sweeps}
\end{figure*}

We perform hyperparameter sweeps of Mistral-7Bv0.3 and Llama3.1-8B using WeightsAndBiases. An exemplary graph of sweep results is shown in Figure \ref{fig:sweeps}. Observing consistently higher performance when using Llama3.1-8B as the base model, we base subsequent experiments on it. The parameters that we vary are shown in Table \ref{tabular:hyperparam_table}. 

\begin{table}[h]
\begin{tabular}{lr}
\hline
\textbf{Hyperparameter}     & \textbf{Search space}                                                                                                                                                \\ \hline
Learning rate               & 4e-5, 1e-4                                                                                                                                                           \\
Learning rate scheduler     & cosine                                                                                                                                                               \\
Optimizer                   & adamw\_torch\_fused                                                                                                                                                  \\
Batch size                  & 2                                                                                                                                                                    \\
Gradient accumulation steps & 8                                                                                                                                                                    \\
LoRA rank                   & 32, 64, 128                                                                                                                                                          \\
LoRA alpha                  & 2 * LoRA rank                                                                                                                                                        \\
LoRA dropout                & 0.0                                                                                                                                                                  \\
LoRA target modules         & \begin{tabular}[c]{@{}r@{}}\{"q\_proj", "v\_proj"\}, \\ \{"q\_proj", "v\_proj", \\ "k\_proj", "o\_proj", \\ "gate\_proj", "down\_proj", \\ "up\_proj"\}\end{tabular} \\
bnb\_4bit\_compute\_dtype   & torch.bfloat16                                                                                                                                                       \\ \hline

\end{tabular}
\caption{The various settings of LoRA and training hyperparameters we experimented with.}
\label{tabular:hyperparam_table}
\end{table}

We find that hyperparameter tweaking only adds a modest 0.5-0.6\% to model scores, with a highest setting at a learning rate of 1e-4, a LoRA rank of 64, and specifying all possible matrices as target modules. 

\end{document}